\documentclass[letterpaper, 10 pt, conference]{ieeeconf}  

\IEEEoverridecommandlockouts                              

\usepackage{graphics} 
\usepackage{epsfig} 
\usepackage{mathptmx} 
\usepackage{times} 
\usepackage{amsmath} 
\usepackage{amssymb}  
\usepackage{algorithm}
\usepackage{algorithmic}
\usepackage{url}
\usepackage{color}
\usepackage{soul}
\usepackage{bm}
\usepackage{subfig}
\usepackage{siunitx}

\graphicspath{{figs/}}

\newcommand{\bA}{\mathbf{A}}

\newcommand{\bB}{\mathbf{B}}
\newcommand{\bb}{\mathbf{b}}
\newcommand{\bC}{\mathbf{C}}

\newcommand{\bc}{\mathbf{c}}
\newcommand{\bD}{\mathbf{D}}

\newcommand{\bI}{\mathbf{I}}

\newcommand{\bu}{\mathbf{u}}

\newcommand{\bP}{\mathbf{P}}

\newcommand{\bR}{\mathbf{R}}
\newcommand{\bS}{\mathbf{S}}

\newcommand{\bt}{\mathbf{t}}

\newcommand{\bX}{\mathbf{X}}

\newcommand{\bZ}{\mathbf{Z}}
\newcommand{\be}{\mathbf{e}}
\newcommand{\bz}{\mathbf{z}}
\newcommand{\cZ}{\mathsf{Z}}
\newcommand{\cX}{\mathsf{X}}
\newcommand{\cN}{\mathsf{N}}
\newcommand{\bK}{\mathbf{K}}

\newcommand{\krot}{known\_rotation\_prob}
\newcommand{\krotd}{krot\_tdc}

\title{\LARGE \bf
Visual SLAM: Why Bundle Adjust?
}

\author{\'{A}lvaro {Parra Bustos}$^{1}$, Tat-Jun Chin$^{1}$, Anders Eriksson$^{2}$ and Ian Reid$^{1}$
\thanks{$^{1}$School of Computer Science, The University of Adelaide.}%
\thanks{$^{2}$School of Electrical Engineering and Computer Science, Queensland University of Technology}%
}

\begin{document}

\maketitle
\thispagestyle{empty}
\pagestyle{empty}

\begin{abstract}

Bundle adjustment plays a vital role in feature-based monocular SLAM. In many modern SLAM pipelines, bundle adjustment is performed to estimate the 6DOF camera trajectory and 3D map (3D point cloud) from the input feature tracks. However, two fundamental weaknesses plague SLAM systems based on bundle adjustment. First, the need to carefully initialise bundle adjustment means that all variables, in particular the map, must be estimated as accurately as possible and maintained over time, which makes the overall algorithm cumbersome. Second, since estimating the 3D structure (which requires sufficient baseline) is inherent in bundle adjustment, the SLAM algorithm will encounter difficulties during periods of slow motion or pure rotational motion.

We propose a different SLAM optimisation core: instead of bundle adjustment, we conduct rotation averaging to incrementally optimise \emph{only camera orientations}. \emph{Given} the orientations, we estimate the camera positions and 3D points via a quasi-convex formulation that can be solved efficiently and \emph{globally optimally}. Our approach not only obviates the need to estimate and maintain the positions and 3D map at keyframe rate (which enables simpler SLAM systems), it is also more capable of handling slow motions or pure rotational motions.

\end{abstract}

\section{INTRODUCTION}

Let $\bu_{i,j}$ be the 2D coordinates of the $i$-th scene point as seen in the $j$-th image $\cZ_j$. Given a set $\{ \bu_{i,j} \}$ of observations, structure-from-motion (SfM) aims to estimate the 3D coordinates $\cX = \{ \bX_i \}$ of the scene points and 6DOF poses $\{ (\bR_j,\bt_j)\}$ of the images $\{ \cZ_j \}$ that agree with the observations. The bundle adjustment (BA) formulation is
\begin{align}\label{eq:ba}
\begin{aligned}
& \underset{\{\bX_i\},\{(\bR_j,\bt_j)\}}{\text{min}}
& & \sum_{i,j} \left\| \bu_{i,j} - f( \bX_i \mid \bR_j, \bt_j )  \right\|_2^2,
\end{aligned}
\end{align}
where $f(\bX_i \mid \bR_j,\bt_j)$ is the projection of $\bX_i$ onto $\cZ_j$ (assuming calibrated cameras). In practice, not all $\bX_i$ are visible in every $\cZ_j$, thus some of the $(i,j)$ terms are dropped. For ease of exposition, we follow~\cite{strasdat12} and regard the image set $\{ \cZ_j \}$ as inputs to BA, bearing in mind that the effective inputs are the observations $\{ \bu_{i,j} \}$ and the visibility matrix.

As a non-linear least squares problem,~\eqref{eq:ba} is usually solved by gradient descent methods, e.g., Levenberg-Marquardt, which require initialisation for all unknowns. Thus, apart from the images $\{ \cZ_j \}$, the total inputs to a BA instance typically include the initial values for $\{(\bR_j,\bt_j)\}$ and $\cX$.

BA is justifiable in the maximum likelihood sense if the errors due to the uncertainty in localising the feature points $\{ \bu_{i,j} \}$ are Normally distributed. However, it is not obvious that available feature detectors satisfy this property~\cite{kanazawa01,kanatani04,kanatani08}. While this does not reduce the usefulness of BA, its statistical validity should not be taken for granted.

\subsection{BA-SLAM}

\begin{algorithm}[t]
\caption{BA-SLAM (adapted from~\cite{strasdat12}).}
\label{alg:ba-slam}
\begin{algorithmic}[1]
\STATE $\cX \leftarrow \text{Initialise\_points}(\cZ_0)$.\label{step:init}
\FOR{each keyframe step $t = 1,2,\dots$}
\STATE $s \leftarrow t - \text{(window size)} + 1$.
\IF{a number of $n \ge 1$ points left field of view}
\STATE $\cX \leftarrow \cX \cup \text{initialise\_}n\text{\_new\_points}(\cZ_t)$.\label{step:spawn}
\ENDIF
\STATE $\bR_{s:t},\bt_{s:t},\cX \leftarrow \text{BA}(\bR_{s:t},\bt_{s:t},\cX,\cZ_{0:t})$.\label{step:window-ba}
\IF{loop is detected in $\cZ_t$}
\STATE $\bR_{1:t},\bt_{1:t},\cX \leftarrow \text{BA}(\bR_{1:t},\bt_{1:t},\cX,\cZ_{0:t})$.\label{step:loop-ba}
\ENDIF
\ENDFOR
\end{algorithmic}
\end{algorithm}

Roughly speaking, monocular feature-based SLAM~\cite{davison07} (henceforth, ``SLAM") is the execution of SfM incrementally to process streaming input images $\cZ_{0:t}$, where
\begin{align}
\cZ_{0:t} = \{ \cZ_0,\cZ_1,\dots,\cZ_t \}.
\end{align}
Several influential works~\cite{engels06,klein07,strasdat12,mur-artal15} have cemented the importance of BA in SLAM. Algorithm~\ref{alg:ba-slam}, which is adapted from~\cite[Table~1]{strasdat12}, describes a SLAM optimisation core based on BA over keyframes. Specifically:
\begin{itemize}
\item In Step~\ref{step:spawn}, new scene points are ``spawned" if the current frame $\cZ_t$ does not adequately observe the map $\cX$.
\item In Step~\ref{step:window-ba} (a.k.a.~\emph{local mappping}), BA is used to estimate the camera trajectory and 3D map in the current time window. Often, local mapping is preceded by \emph{camera tracking} to accurately initialise the current pose $(\bR_t,\bt_t)$. See~\cite[Sec.~5.3]{strasdat12} or~\cite[Sec.~V]{mur-artal15} for examples.
\item In Step~\ref{step:loop-ba} (a.k.a.~\emph{loop closure}), a system-wide BA is executed to reoptimise all the variables and redistribute accumulated drift errors. Implicit in Algorithm~\ref{alg:ba-slam} is the introduction of covisibility information between $\cZ_t$ and older keyframes, prior to BA. Often, Step~\ref{step:loop-ba} is preceded by \emph{pose graph optimisation}~\cite{grisetti09,kummerle11,kaess12,carlone15} to give a more accurate initialisation of the poses.
\end{itemize}
Note that Algorithm~\ref{alg:ba-slam} is merely a ``basic recipe" for SLAM. In practice, ``\emph{what will make or break a real-time SLAM system are all the (often heuristic) nitty-gritty details}"~\cite{engel17}, e.g., how to select features/keyframes, how to update the covisibility graph, how to select/merge/prune 3D points, etc. However, since our focus is on optimisation, Algorithm~\ref{alg:ba-slam} is sufficient to capture the core algorithmic elements of SLAM systems based on BA, such as ORB-SLAM~\cite{mur-artal15}.

\subsection{Why do we want an alternative to BA-SLAM?}

\subsubsection{High system complexity}

Besides computing the trajectory and map in the current vicinity, Step~\ref{step:window-ba} in BA-SLAM plays a more basic role: incrementally estimating the variables in small BA ``chunks" to serve as initialisation for the system-wide BA in Step~\ref{step:loop-ba}. Note that since~\eqref{eq:ba} is amenable to only locally optimal solutions, without good initial values for the large number of variables (poses and 3D points), Step~\ref{step:loop-ba} will converge to poor solutions.

Therefore, unavoidably all variables must be estimated as accurately as possible and updated at keyframe rate throughout the lifetime of the algorithm---we argue that this increases the complexity of SLAM systems. For example, while Algorithm~\ref{alg:ba-slam} shows only the creation of new 3D points (Step~\ref{step:spawn}), in a practical system (e.g.,~\cite{mur-artal15}) a host of other heuristics are required for map maintenance, e.g., map point selection, point culling, map updating and aggregation. Many of these heuristics contain a number of thresholds, which, if not tuned carefully, will lead to system failure.



\subsubsection{Difficulties due to pure rotational motion}

More fundamentally, since the estimation of 3D points (which require sufficient baseline) are essential, it is unavoidable that a system based on BA-SLAM will encounter numerical issues during periods of pure rotational motion or slow motion~\cite[Sec.~7.1]{taketomi17}, and will require special treatment to deal with this problem~\cite{gauglitz12,pirchheim13,herrera14,tang17}. This issue is particularly acute at the start of the sequence where the camera is usually slow moving\footnote{Pioneers of monocular SLAM~\cite{davison07} call the deliberate motion to initialise a SLAM algorithm the ``SLAM wiggle".}. For example, in ORB-SLAM, elaborate map initialisation heuristics~\cite[Sec.~IV]{mur-artal15} (cf.~Step~\ref{step:init} in Algorithm~\ref{alg:ba-slam}) are used to combat inaccuracies due to insufficient translations. However, experienced users of ORB-SLAM cite its difficulty to initialise on challenging image sequences.

\section{L-infinity SLAM}\label{sec:linf-slam}

Towards simpler visual SLAM systems, we propose a novel optimisation core called \emph{L-infinity SLAM}; see Algorithm~\ref{alg:linf-slam}. A main distinguishing factor is that the online effort (Steps~\ref{step:window-rotavg} and~\ref{step:loop-rotavg}) are devoted to estimating only the camera orientations via \emph{rotation averaging}~\cite{hartley13,eriksson18}. Given the orientations, a separate optimisation via the \emph{known rotation problem}~\cite{kahl05,olsson07} (Steps~\ref{step:window-krot} and~\ref{step:loop-krot}) is conducted to obtain the camera positions and 3D map---since rotation averaging can be done independently from position and map estimation, Steps~\ref{step:window-krot} and~\ref{step:loop-krot} can be performed in a lower priority thread.

\begin{algorithm}[ht]
\caption{L-infinity SLAM.}
\label{alg:linf-slam}
\begin{algorithmic}[1]
\FOR{each keyframe step $t = 1,2,\dots$}
\STATE $s \leftarrow t-(\text{window size}) + 1$.
\STATE $\{ \bR_{j,k} \}_{j, k \in \cN_{\text{win}}} \leftarrow \text{relative\_rotation}(\cZ_{(s-1):t})$.\label{step:relrot}
\STATE $\bR_{s:t} \leftarrow \text{rotation\_averaging}\left(   \{\bR_{j,k} \}_{j,k \in \cN_{\text{win}}} \right)$.\label{step:window-rotavg}
\STATE $\bt_{s:t},\cX \leftarrow \text{\krot{}}(\bR_{s:t},\cZ_{0:t})$.\label{step:window-krot}
\IF{loop is detected in $\cZ_t$}
\STATE $\{ \bR_{j,k} \}_{j, k \in \cN_{\text{sys}}} \leftarrow \text{relative\_rotation}(\cZ_{0:t})$.\label{step:relrot2}
\STATE $\bR_{1:t} \leftarrow \text{rotation\_averaging}\left(\{ \bR_{j,k} \}_{j,k \in \cN_{\text{sys}}}\right)$.\label{step:loop-rotavg}
\STATE $\bt_{1:t},\cX \leftarrow \text{\krot{}}(\bR_{1:t},\cZ_{0:t})$.\label{step:loop-krot}
\ENDIF
\ENDFOR
\end{algorithmic}
\end{algorithm}

\subsection{Rotation averaging and known rotation problem}

Given a set of relative rotations $\{ \bR_{j,k} \}$ between pairs of overlapping images $\{\cZ_j,\cZ_k\}$, the goal of rotation averaging is to estimate the absolute rotations $\{ \bR_j \}$ that are consistent with the relative rotations (Sec.~\ref{sec:relmotions} will provide details on estimating relative rotations in our work). Following~\cite{eriksson18}, we chose the chordal metric for rotations, which yields the rotation averaging formulation
\begin{align}\label{eq:rotavg}
\begin{aligned}
& \underset{\{\bR_j \}}{\text{min}}
& & \sum_{j,k \in \cN} \left\| \bR_{j,k} - \bR_k \bR_j^{-1} \right\|_F^2,
\end{aligned}
\end{align}
where $\cN$ is the covisibility graph. In Step~\ref{step:relrot} of L-infinity SLAM, the covisibility graph $\cN_{\text{win}}$ in the window is used, while in Step~\ref{step:loop-rotavg}, the system-wide covisibility graph $\cN_{\text{sys}}$ (updated to account for loop closure) is used. Sec.~\ref{sec:algo} will describe the specific algorithm for~\eqref{eq:rotavg}.

Given a set of absolute camera orientations $\{ \bR_j \}$, the known rotation problem (KRot)~\cite{kahl05} optimises the camera positions $\{ \bt_j \}$ and 3D points $\{ \bX_i \}$ as
\begin{align}\label{eq:krot}
\begin{aligned}
& \underset{\{ \bX_i \}, \{ \bt_j \}}{\text{min}}
& & \max_{i,j}~~\| \bz_{i,j} -  f( \bX_i \mid \bR_j, \bt_j ) \|_2,
\end{aligned}
\end{align}
subject to cheirality constraints (details in Sec.~\ref{sec:algo}). Observe that unlike~\eqref{eq:ba} which minimises the sum of squared reprojection errors,~\eqref{eq:krot} minimises the maximum reprojection error, which can be viewed as the $\ell_\infty$-norm of the vector of reprojection errors (leading to the name ``L-infinity SLAM").

At this juncture, it is vital to note that~\eqref{eq:krot} is quasi-convex, which is amenable to \emph{efficient global solution}~\cite{kahl05,olsson07}. In our work, a novel variant of KRot is proposed specifically for the loop closure optimisation in Step~\ref{step:loop-krot}; details in Sec.~\ref{sec:algo}.

\subsection{Benefits of L-infinity SLAM}

\subsubsection{Simplicity}

As alluded to above, tracking and loop-closing in L-infinity SLAM estimate only orientations. Since positions and 3D map are obtained via an independent optimisation problem that can be solved globally optimally, the results of Steps~\ref{step:window-krot} and~\ref{step:loop-krot} do not affect the results of subsequent instances\footnote{Their results can be used to warm start the subsequent instances, but this is an optional computational consideration.}. Therefore, there is no need to accurately calculate positions and 3D map on-the-fly and maintain/propagate them. Note that in Algorithm~\ref{alg:linf-slam}, Steps~\ref{step:window-krot} and~\ref{step:loop-krot} are shown mainly to make the overall functionality of L-infinity SLAM equivalent to BA-SLAM. Contrast this to the equivalent steps in BA-SLAM (Steps~\ref{step:window-ba} and~\ref{step:loop-ba} in Algorithm~\ref{alg:ba-slam}), whose resulting quality are vital at all times to ensure correct operation.

A signficant advantage of the processing flow of L-infinity SLAM is that many tasks related to map maintenance (e.g., feature/map point selection, point culling, map updating and aggregation) can be done in a low priority thread, or even offline if there is no need for on-the-fly position and map estimation (e.g., the application in~\cite{khosravian17}). This has the potential to significantly simplify visual SLAM systems.



\subsubsection{Handling pure rotation motion}

It is well-known that under epipolar geometry, camera orientation can be estimated independently from the translation~\cite{kneip14}. Hence, since the online routines in L-infinity SLAM estimate orientations only, a real-time system based on L-infinity SLAM is less likely to encounter difficulties due to pure or close-to-pure rotational motions. Potential numerical issues due to insufficient baselines between camera views can be handled in the low-priority thread that estimates position and 3D map. Sec.~\ref{sec:lowspeed} will provide results that illustrate this advantage of L-infinity SLAM over BA-SLAM.



\subsection{Concerns on global optimality and outliers}


Some readers may find it disconcerting that in L-infinity SLAM the estimation of the variables are detached. \emph{First, note that we have not claimed that L-infinity SLAM is globally optimal in all variables.} Second, there is ample evidence~\cite{carlone15b,eriksson18} that rotation averaging algorithms are capable of producing highly-accurate orientation estimates, independently from positions and 3D points. Since the quasiconvex estimation for positions and 3D points is globally optimal, the overall quality of L-infinity SLAM will be high, as we will demonstrate in Sec.~\ref{sec:results}.


Also, a common impression of $\ell_\infty$ estimation is its sensitivity to outliers. Note, however, that both the $\ell_\infty$ and $\ell_2$ norms have a breakdown point of 0~\cite{rousseeuw87}, hence both norms are equally susceptible to outliers. In practical BA-SLAM systems, a typical remedy is to pass the $\ell_2$ residual through an isotropic robust norm (e.g., Cauchy norm). Likewise, there are efficient and theoretically justified techniques to identify and remove outliers in $\ell_\infty$ estimation~\cite{ke07,sim06out}. Hence, outliers do not present a problem for L-infinity SLAM.

\section{Algorithmic details}\label{sec:algo}

In this section, we describe the details of the core optimisation routines in L-infinity SLAM. Consider a calibrated camera with $\bK$ the $\mathbb{R}^{3 \times 3}$ camera intrinsic matrix. Let 
\begin{align}
\bP_j = \bK \left[ \bR_j~\bt_j \right]\label{eq:proj}
\end{align}
be the projection matrix of the j-th image with assumed known rotation matrix $\bR_j$ in $SO(3)$ and unknown translation vector $\bt_j$ in $\mathbb{R}^{3}$. For simplicity, we assume $\bK = \bI_{3 \times 3}$. For an arbitrary $\bK$, derivations are still valid if camera extrinsics are recovered by applying $\bK^{-1}$ to $\bR_j$ and $\bt_j$, which now form the three first columns and the last column of $\bP_j$.

\subsection{Estimating relative motions}\label{sec:relmotions}

L-infinity SLAM estimates camera rotations from relative camera rotations $\bR_{j,k}$ in the covisibility graph $\cN_{\text{win}}$. We simply estimate $\bR_{j,k}$ from the essential matrix which can be decomposed into $\bR_{j,k}$ and a relative translation direction $\bt_{j,k}^{(E)}$ ($\|\bt_{j,k}^{(E)}\|=1$). A weakness of this decomposition is the need of images with sufficient displacement; however, other methods can be used to estimate relative motions~\cite{kneip14,ventura15,ha18}. In the case of low displacement, we estimated $\bR_{j,k}$ by rotationally aligning backprojected feature rays by using a rotation only variant of Trimmed ICP~\cite{chetverikov02}. 

\subsection{Rotation averaging}
Several methods exist to solve~\eqref{eq:rotavg}~\cite{hartley11,hartley13,chatterjee13,eriksson18}. Here we adopt the robust method of~\cite{chatterjee13} which uses an iteratively reweighted least-squares approach in $SO(3)$. The method in~\cite{chatterjee13} is simpler than BA as, for example, no linearisation and no estimation of a damping factor is required.
 


\subsection{Known rotation problem}
By referring to $\bR_j^{(1:2)}$ as the first two rows of $\bR_j$, and to $\bR_j^{(3)}$ as the third row of $\bR_j$ (similarly for $\bt^{(1:2)}$ and $\bt^{(3)}$), the projection of $\bX_i$ onto the $j$-th image is given by
\begin{align}
f(\bX_i \mid \bR_j,\bt_j) :=\frac{\bR^{(1:2)}_j\bX_i + \bt_j^{(1:2)}} {\bR^{(3)}_j \bX_i + \bt_j^{(3)}},
\end{align}
and the known rotation problem~\eqref{eq:krot} can be rewritten by adding an extra variable $\gamma$ as
\begin{subequations}
\begin{align}
&&P_{0}:&&\underset{\{ \bX_i \}, \{ \bt_j \}}\min &\quad\gamma \\
&&&&\text{subject to}\quad &  \dfrac
{\left\| \bA_{i,j} \begin{bmatrix} \bX_i \\ \bt_j \end{bmatrix} \right\|_2 }
{\bb_j^{\top} \begin{bmatrix} \bX_i \\ \bt_j \end{bmatrix} }
 \leq \gamma,   \quad \forall \, i,j,\label{const:levelset}\\
&&&& &  \bb_j^{\top} \begin{bmatrix} \bX_i \\ \bt_j \end{bmatrix}   \geq 0,  \quad \forall \, i,j, &\label{const:cheirality}\\
&&&& & \gamma \geq 0,
\end{align}
\end{subequations}
where 
\begin{align}
&\bA_{i,j} = \begin{bmatrix} \bS_{i,j} \;\; \bI_{2 \times 2} \;\; -\bz_{i,j}
\end{bmatrix}, \quad
\bb_{j}  = \left[ \bR_j^{(3)} \;0\; \;0\; \;1 \right]^\top, \quad \text{and}\\
&\bS_{i,j} = \bR_j^{(1:2)}-\bz_{i,j}\bR_j^{(3)}.
\end{align}

Cheirality constraints~\eqref{const:cheirality} impose to $\bX_i$ to lie in front of the cameras in which $\bX_i$ is visible. 

Intuitively, $\gamma$ defines the sublevel sets of the objective in~{\eqref{eq:krot}}, i.e., the maximum over the LHS of~{\eqref{const:levelset}}. For a fixed $\gamma$,
\begin{align}
\left\| \bA_{i,j} \begin{bmatrix} \bX_i \\ \bt_j \end{bmatrix} \right\|_2 - \gamma \; \bb_j^{\top} \begin{bmatrix} \bX_i \\ \bt_j \end{bmatrix} \;\leq \;0
\end{align}
defines a convex set hence the objective is quasi-convex and {$(P_0)$} is a quasi-convex problem. For a detailed proof of $(P_0)$ being a quasi-convex problem the reader can refer to~\cite{kahl08}.

\subsubsection{Solving the known rotation problem}
$(P_{0})$ can be rewritten as
\begin{subequations}	
\begin{align}
&P_{1} : & \underset{\{ \bX_i \}, \{ \bt_j \}}\min &\quad\gamma& \\
&& \text{subject to} \quad&  \left\| \bA_{i,j} 
\begin{bmatrix} \bX_i\\ 
\bt_j \end{bmatrix}  \right\|_2 \leq \gamma \; \bb_j^{\top} \begin{bmatrix} \bX_i \\ 
\bt_j \end{bmatrix},  \; \forall \, i,j, \label{const:socp}\\
&& & \gamma \geq 0,
\end{align}
\end{subequations}
in which the Cheirality constraints~\eqref{const:cheirality} are implicit in~\eqref{const:socp} as both the LHS of~\eqref{const:socp} and $\gamma$ are non-negative. If $\gamma$ is fixed, constraints~\eqref{const:socp} became second-order cones which allows to solve ($P_{1}$) by using bisection trough SOCP feasibility tests~\cite{kahl05}. Several other methods solve $(P_1)$ through SOCP sub-problems. \cite{agarwal08} shown that Gugat's algorithm~\cite{gugat96} outperforms among this type of methods (bisection, Brent's algorithm\cite{brent13}, and Dinkelbach's algorithm~\cite{dinkelbach67,olsson07}). Recently, Zhang et al.~\cite{zhang18} presented a method, named Res-Int, which outperformed existent methods by alternating between pose estimation and triangulation to efficiently partition the problem into small sub-problems without compromising global optimality. As a result, Res-Int solves $(P_1)$ in about 3 seconds for moderate size input (around 15 images and 3000 3D points). 

We use Res-Int as the \krot{} routine in Line~\ref{step:window-krot} in Algorithm~\ref{alg:linf-slam} since its superior performance. Although Res-Int converges in a few seconds for moderate problems (as those optimised for a moving window in Algorithm~\ref{alg:linf-slam}), its performance is still inadequate for medium to large size problems ($>10,000$ 3D points, $>100$ images) that Algorithm~\ref{alg:linf-slam} optimises when detecting a loop. For such problems, Res-Int could take from a few minutes to hours to converge (see~\cite[Table~2]{zhang18}).  To efficiently address loop closure, we propose a new formulation that incorporates relative camera translation directions (obtained from the essential matrix) to alleviate the size of the problem but still produce an accurate result. 

\subsection{Known rotation problem with translation direction constraints}
Solving KRot in Step~\ref{step:loop-krot} in Algorithm~\ref{alg:linf-slam} can be excessively time-consuming as a loop can be detected at an advanced stage generating a large input size. Instead, we propose to address loop closure over a sample of the input with a formulation that incorporates camera translation directions.

Inspired by the quasi-convex approach of Sim and Hartley~\cite{sim06} to estimate the camera translations from $\bt_{j,k}^{(E)}$ and known camera rotations, we constrained camera positions
\begin{align}
\bC_j = -\bR_j^\top \bt_j
\end{align}
in the known rotation problem $(P_1)$ to agree up to an angular threshold 
\begin{align}
\angle(\bt_{j,k}, \bC_k - \bC_j) \leq \alpha \quad \forall j,k,\label{eq:angconst}
\end{align}
to 
\begin{align}
\bt_{j,k} = (\bK^{-1}\bR_j)^{\top}\bt_{j,k}^{(E)}\label{eq:t_jk},
\end{align}
which is the relative translation direction in world coordinates (we explicitly apply $\bK^{-1}$ to $\bR_j$ in~\eqref{eq:t_jk} in case $\bK \neq \bI_{3 \times 3}$ and therefore $\bR_j$ is not a rotation matrix).

We observed that the method of~\cite{sim06} was unable to produce satisfactory results (see results in Sec.~\ref{sec:results}) for loop closure, arguably since no structural information is optimised thus camera positions were not sufficiently constrained. On the other hand, adding angle constraints~\eqref{eq:angconst} to $(P_1)$ it allows to efficiently solve loop closure ($<100$~s) with a sparse set of 3D points (300 points for result in Fig.~\ref{fig:superposition_maptek_linf}). Our proposition yields the following problem.
\begin{subequations}	
	\begin{align}
	&P_{3} : & \underset{\{ \bX_i \}, \{ \bC_j \}}\min &\quad\gamma& \\
	&& \text{subject to} \quad&  \left\| \bB_{i,j} 
	\begin{bmatrix} \bX_i\\ 
	\bC_j \end{bmatrix}  \right\|_2 \leq \gamma \;\bc_{j}^{\top} \begin{bmatrix} \bX_i \\ \bC_j \end{bmatrix},  \; \forall \, i,j, \\
	&& \quad&  \left\| \bD_{j,k} 
	\begin{bmatrix} \bC_j\\ 
	\bC_k \end{bmatrix}  \right\|_2 \leq \be_{j,k}^{\top} \begin{bmatrix} \bC_j \\ \bC_k \end{bmatrix},  \; \forall \, j,k, \label{const:tdirs}\\
	&& & \gamma \geq 0,
	\end{align}
\end{subequations}
where
\begin{align}
&\bB_{i,j} = \begin{bmatrix} \bS_{i,j}\;  -\bS_{i,j}\end{bmatrix}, 
&&\bc_{j} = \begin{bmatrix} \bR_j^{(3)}\; -\bR_j^{(3)}\end{bmatrix}, \\
&\bD_{j,k} = \begin{bmatrix}\bZ_{j,k}^{(1:2)}\; -\bZ_{j,k}^{(1:2)} \end{bmatrix}, 
&&\be_{j,k} = \tan(\alpha) \begin{bmatrix}\bt_{j,k}^{\top}\; -\bt_{j,k}^{\top}\end{bmatrix}^{\top},
\end{align}
and $\bZ_{j,k}$ is a rotation matrix such that 
\begin{align}
\bZ_{j,k} \bt_{j,k} = [0\; 0\; 1]^{\top}.
\end{align}

Similarly to the method in~\cite{sim06}, $(P_3)$ is valid for $\alpha<\ang{90}$. For derivation details of the camera translation direction constraints~\eqref{const:tdirs} please refer to~\cite{sim06}.

\section{Results}\label{sec:results}
Here we compare L-infinity SLAM (Algorithm~\ref{alg:linf-slam}) against BA-SLAM (Algorithm~\ref{alg:ba-slam}) on real data with precise ground truth. The used dataset, provided by Maptek\footnote{\url{https://www.maptek.com/}}, was captured with a system  equipped with a high precision INS (refer to~\cite{khosravian17} for system's details). Mounted on a truck, a forward looking camera captured a video together with inertial measurements providing the ground truth for the camera poses. 


Experiments were run on a PC with a quad-core 2.5GHz Intel core i7 CPU and 16GB of RAM. We implemented L-infinity SLAM and BA-SLAM in MATLAB with the following optimisation routines:
\begin{itemize}
	\item BA: implemented in C++ using the Ceres solver~\cite{ceres-manual}. 
	\item rotation\_averaging: code provided in~\cite{chatterjee13}.
	\item \krot{}: code provided in~\cite{zhang18}.
	\item \krotd{}: $(P_3)$ implemented in MATLAB using \mbox{SeDuMi}~\cite{sturm99}.
\end{itemize}

\subsection{Results for the Maptek dataset}
\begin{figure}
	\subfloat[]{
		\includegraphics[width=\linewidth]{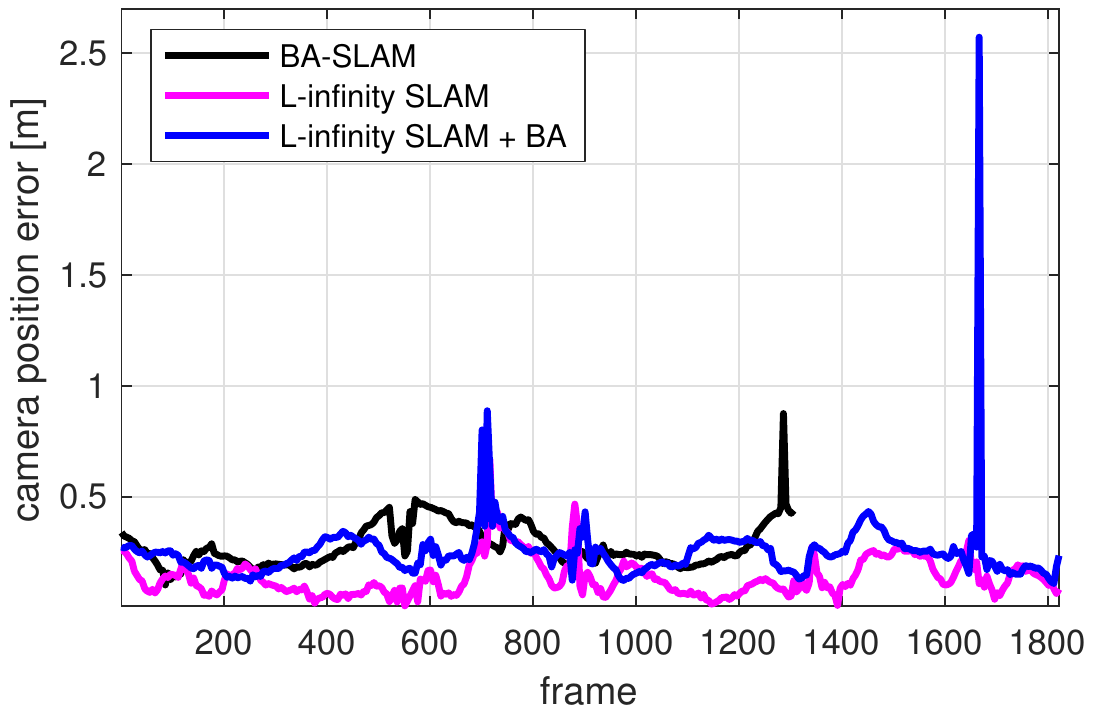}\label{fig:trerr_maptek}}
	
	\subfloat[]{
		\includegraphics[width=\linewidth]{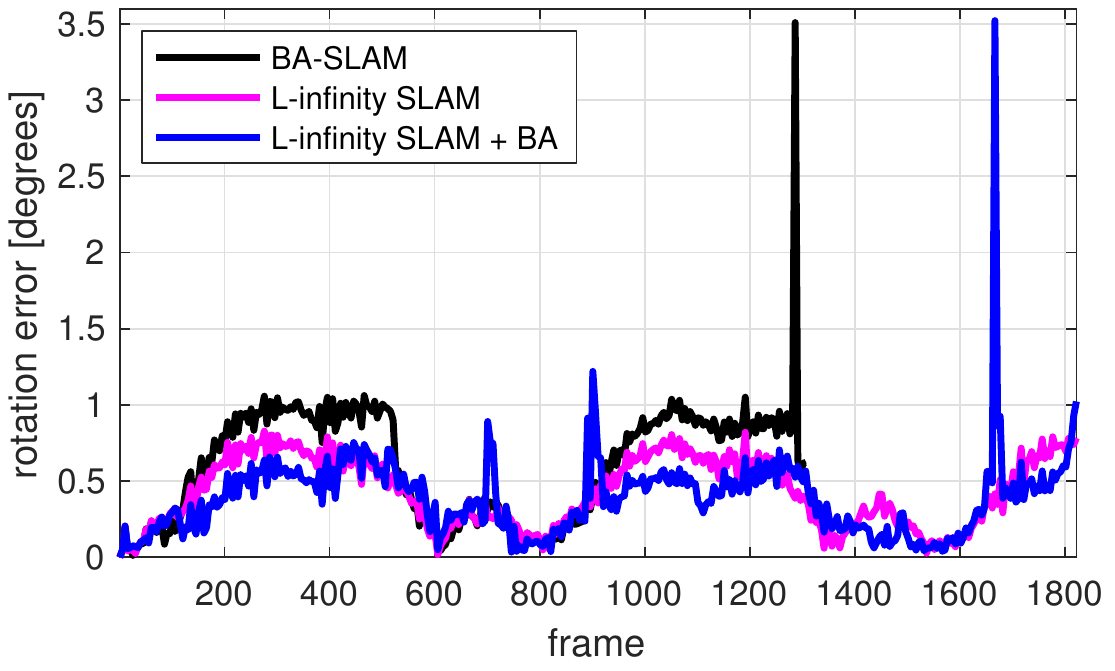}}
	\caption{(a) Camera position error, and (b) camera rotation error for BA-SLAM and L-infinity SLAM in the Maptek dataset. The comparison includes the result of BA after \krotd{} fed with same feature tracks. }
	\label{fig:err_maptek}
\end{figure}

\begin{figure}
	\centering
	\subfloat[BA-SLAM]{
		\includegraphics[width=.85\linewidth]{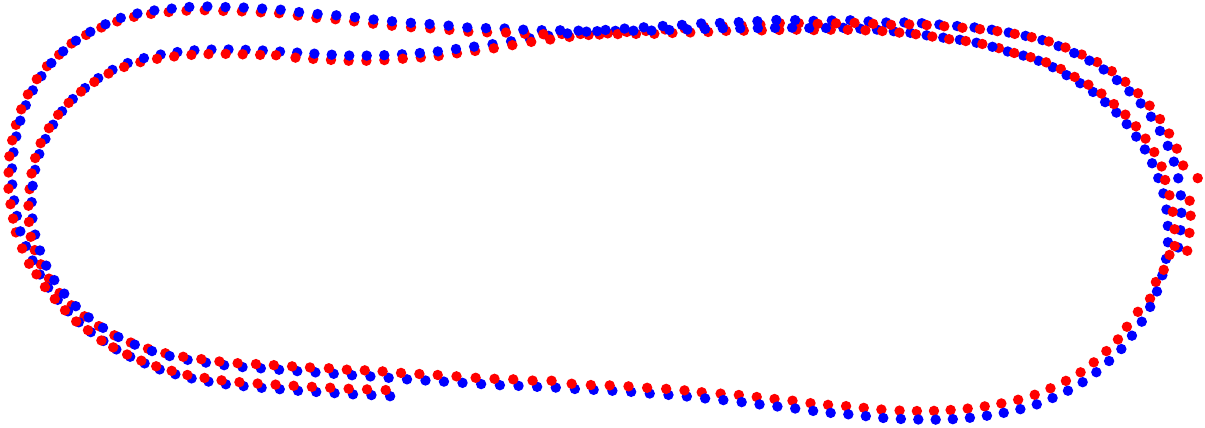}}
	
	\subfloat[L-infinity SLAM with \krotd{}]{
		\includegraphics[width=.85\linewidth]{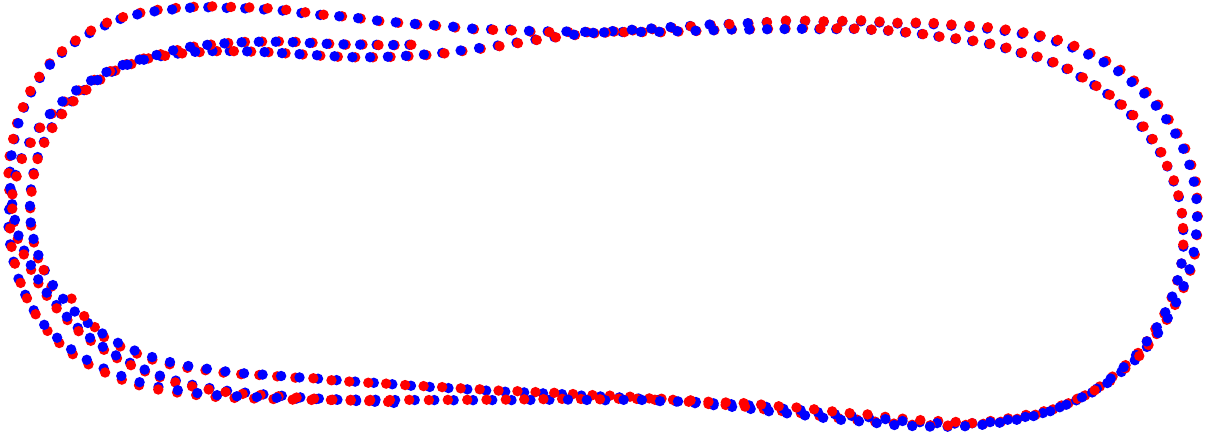}\label{fig:superposition_maptek_linf}}
	\caption{Estimated camera positions (red dots) superposed with the ground truth (blue dots) for (a) BA-SLAM, and (b) L-infinity SLAM solving loop closure with the proposed \krotd{}.}
	\label{fig:superposition_maptek}
\end{figure}

We sampled the full sequence (1833 frames) into 358 keyframes. We detected the occurrence of a loop by using provided ground truth in both BA-SLAM and L-infinity SLAM. Since the moving camera describes a two-loop sequence (see the ground truth in Fig.~\ref{fig:superposition_maptek}), after completing the first loop (at frame 790), a loop is detected for each consecutive keyframe. To solve loop closure in L-infinity SLAM, we fed \krotd{} with 300 uniformly sampled feature tracks (we used the same sample size for loop closure in BA-SLAM). \krotd{} accurately solved loop closure (see Fig~\ref{fig:err_maptek}) in 90.54 s in a MATLAB single-thread implementation. Since its quasi-convex nature, \krotd{} does not have to be invoked at each loop detection; here we invoked \krotd{} at the last keyframe only. On the contrary, BA-SLAM could be incapable of fixing the drift produced if invocations of BA are skipped (e.g., failing in detecting a loop on a real-world system). We set a window size equal to 10.



To compare BA-SLAM and L-infinity SLAM, Fig.~\ref{fig:err_maptek} plots the camera position error and camera rotation error for both methods. BA-SLAM was unable to complete the sequence --the camera got disconnected at frame 1356, i.e., no feature track existed for the visited keyframe. This disconnection was a consequence of the outlier removal heuristic used for BA-SLAM (removing a feature track if the distance of the camera position to any visible 3D point is above a threshold; 150~m for this experiment) which failed by eliminating inliers when BA was unable to produce an accurate result. We observed that BA is prone to fail on reduced data input, hence BA-SLAM needs to keep a large number of feature tracks. The $l_\infty$ optimisation approach of L-infinity SLAM admits less risky outlier removal strategies (e.g. eliminating the support set) without the need of keeping a large number of tracks. As result, BA-SLAM achieved lower camera position ($<0.67~m$) and camera rotation ($<\ang{0.83}$) errors than BA-SLAM. 


We also ran BA after \krotd{} on same feature tracks. The camera position and rotation error tend to be lowered; however, BA produced significant error in several camera poses as depicted in Fig.~\ref{fig:err_maptek}.

\subsubsection{\krotd{} vs the camera recovering method in~\cite{sim06}}
\begin{figure}
	\centering
	\includegraphics[width=.9\linewidth]{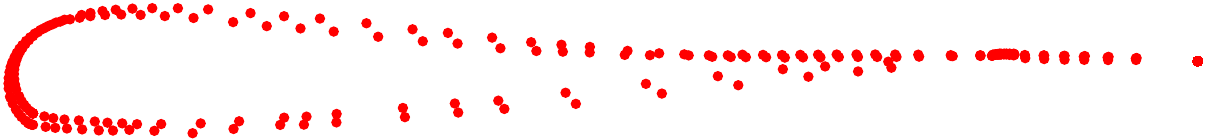}
	\caption{Camera positions obtained with method in~\cite{sim06}.}
	\label{fig:sim}
\end{figure}

\begin{figure}
	\centering
	\includegraphics[width=.85\linewidth]{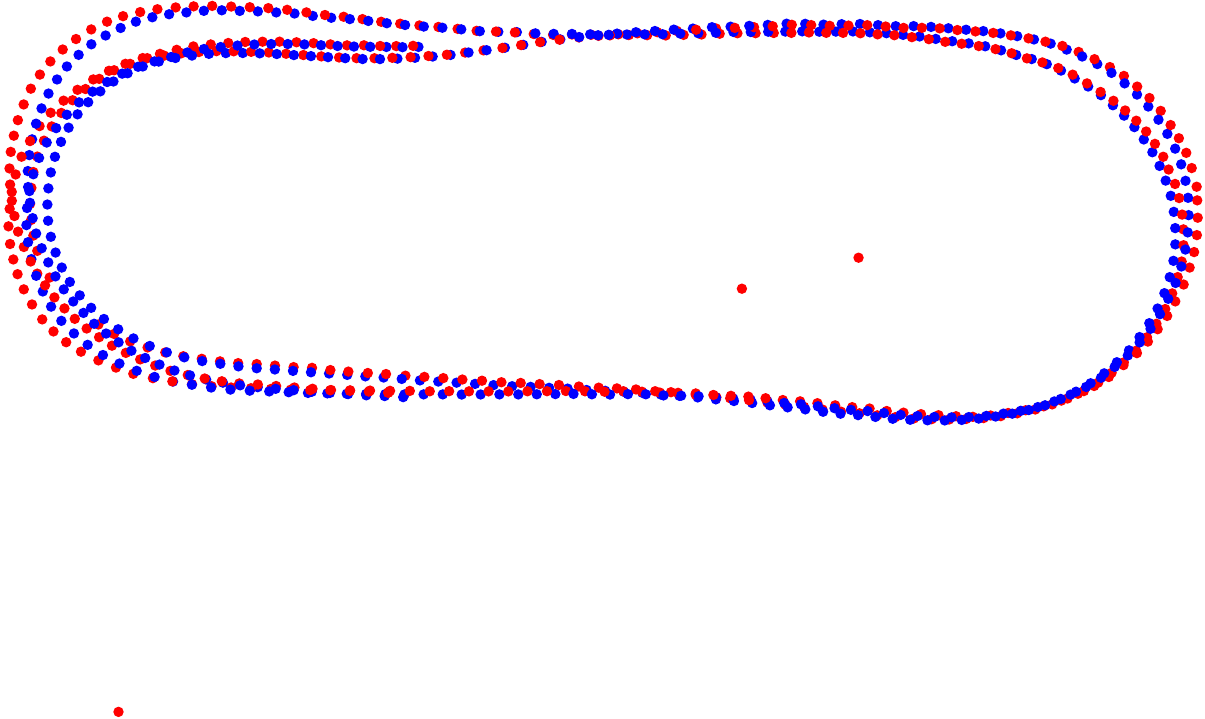}
	\caption{Superposition of \krot{} camera positions (red dots) with the ground truth (blue dots). }
	\label{fig:krot_lc}
\end{figure}

We ran the method in~\cite{sim06} with the same relative translation directions $\bt_{j,k}^{(E)}$ and camera rotations used to solve loop closure. As depicted in Fig.~\ref{fig:sim}, recovering camera positions is unachievable from  $\bt_{j,k}^{(E)}$ measurements only. In addition, Fig.~\ref{fig:krot_lc} shows \krot{} failed on producing an accurate result when solving loop closure on the same tracks we used with \krotd{}.

\begin{figure*}[t!]
	\subfloat[Input]{
		\begin{tabular}[b]{c}
			\includegraphics[width=.24\linewidth]{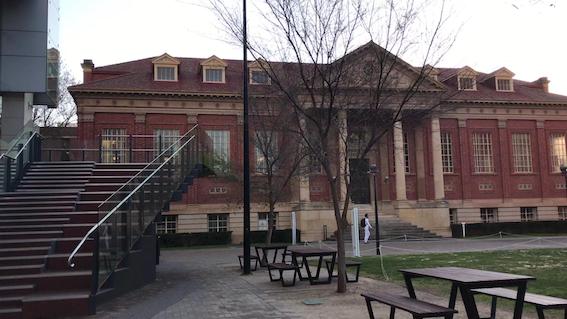}\\[.5em]
			\includegraphics[width=.24\linewidth]{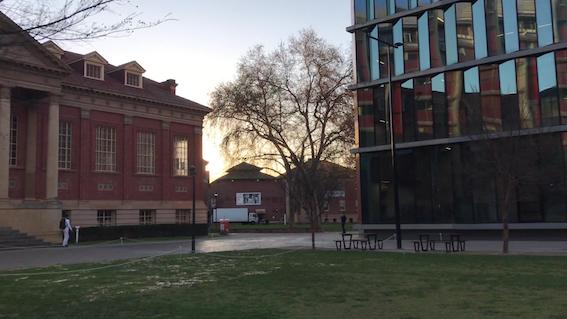}
		\end{tabular}\label{fig:slow_input}
	}
	\subfloat[L-infinity SLAM]{
		\includegraphics[trim={.5cm 0cm 2.5cm 6.9cm}, clip, width=.28\linewidth]{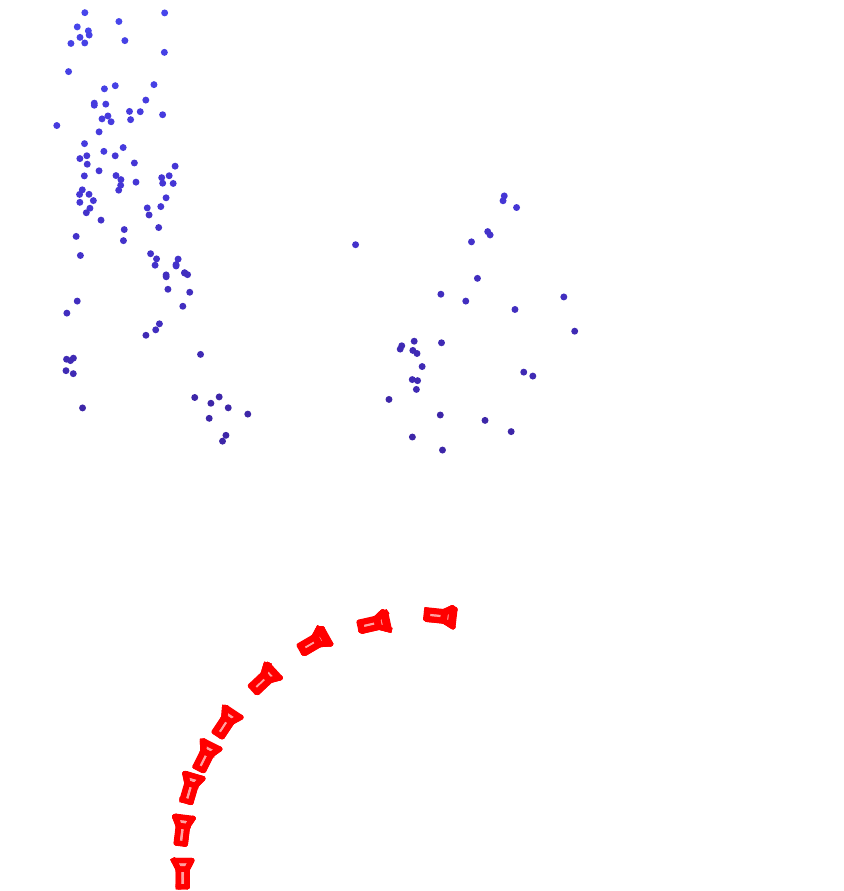}\label{fig:slow_linf}
	}\hspace{-1cm}
	\subfloat[ORB-SLAM]{
		\begin{tabular}[b]{l}
			\includegraphics[width=.22\linewidth]{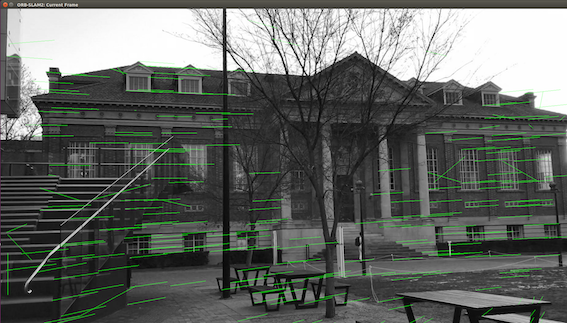}\;\;\includegraphics[width=.22\linewidth]{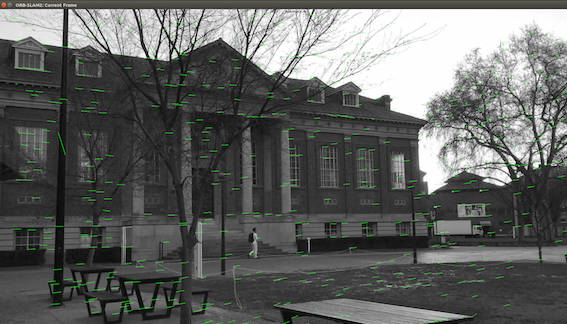}\\[.5em]
			\includegraphics[width=.22\linewidth]{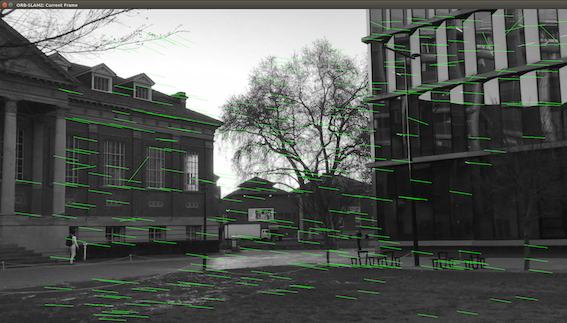}\;\;\includegraphics[width=.22\linewidth]{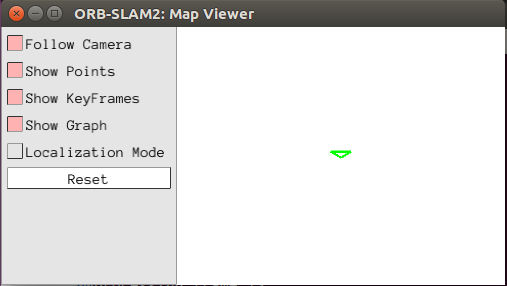}
		\end{tabular}\label{fig:slow_orbslam}
	}
	\caption{A pedestrian recorded a scene while walking and rotating a smart-phone camera. (a) Frame samples of the input video. (b) L-infinity SLAM scene reconstruction. (c) ORB-SLAM failed to initialise hence no reconstruction was possible. Green lines indicate unsuccessful initialisation. The bottom right screenshot displays result (blank) at the end of the sequence.}
	\label{fig:slow}
\end{figure*}

\subsubsection{Map reconstruction}
\begin{figure}
	\includegraphics[trim={2cm 2.5cm 2cm 3cm}, clip, width=\linewidth]{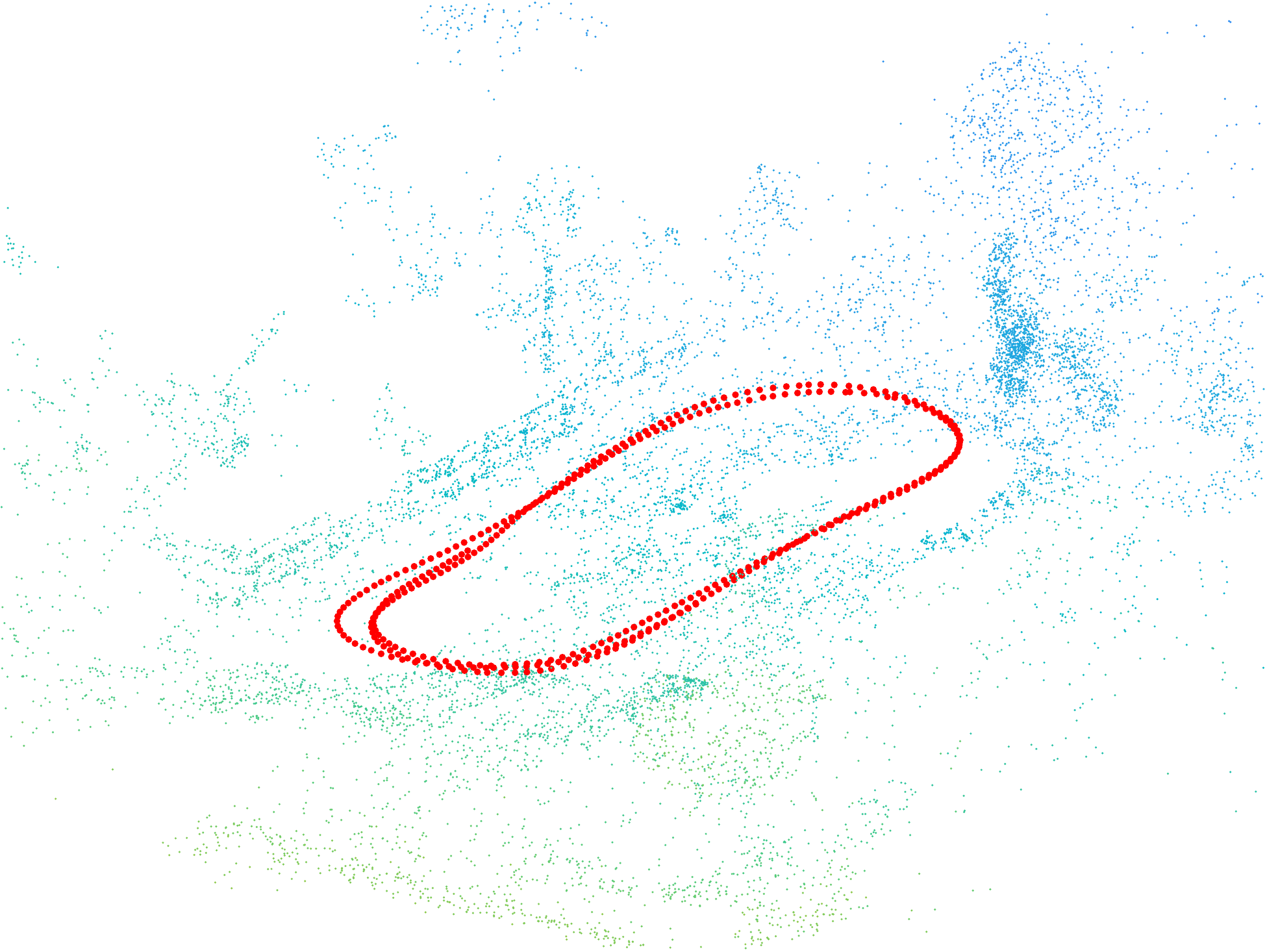}
	\caption{L-infinity SLAM reconstruction.}
	\label{fig:linf_rec}
\end{figure}

Since we reconstructed 300 scene points only when solving loop closure, we used the quasi-convex method in~\cite{zhang18} to triangulate all scene points. Fig.~\ref{fig:linf_rec} shows the scene reconstruction and the camera positions. 

\subsection{Low speed and high rotational motion}\label{sec:lowspeed}
We tested L-Infinity SLAM against ORB-SLAM with a video recorded with a smart-phone by a pedestrian while turning right; Fig.~\ref{fig:slow_input} displays two frames of the video. ORB-SLAM failed to build an initial map (see Fig.~\ref{fig:slow_orbslam}), arguably, since the low baseline of frames from the walking speed sequence with rotational motion (see camera poses in Fig.~\ref{fig:slow_linf}). Unlike ORB-SLAM, the quasi-convex formulation of L-Infinity SLAM does not required any initial map to track camera motions. As result, only L-infinity SLAM produced a reconstruction (see Fig.~\ref{fig:slow_linf}). 

\begin{figure}
	\centering
	\includegraphics[width=\linewidth]{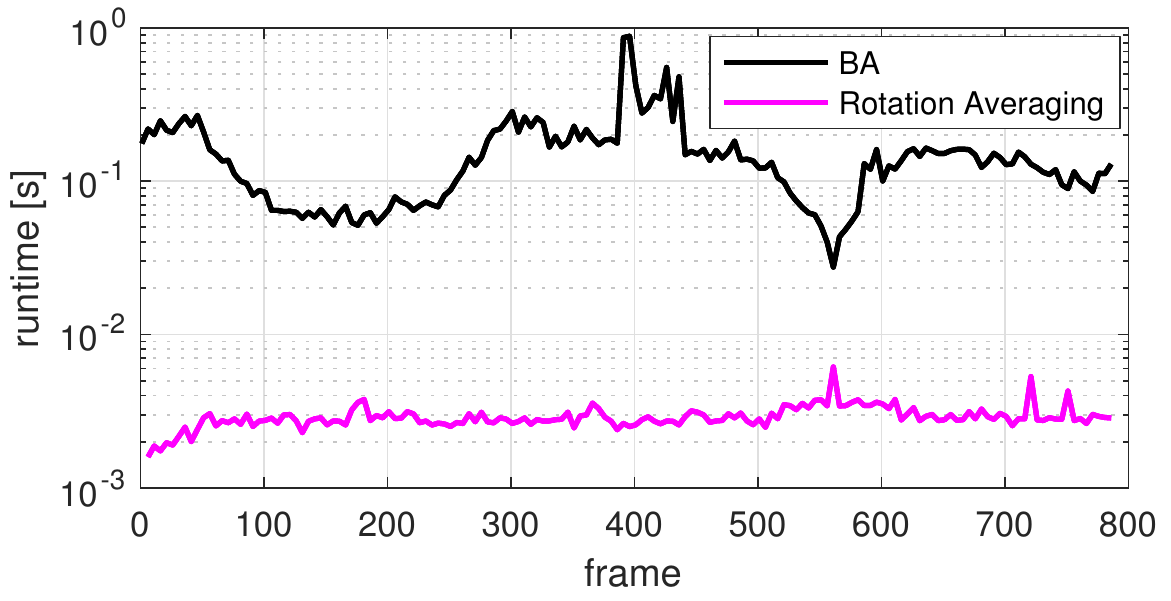}
	\caption{Runtime comparison of incremental rotation averaging and BA.}
	\label{fig:rotonly}
\end{figure}

\subsection{Runtime of online routines}
To compare the efficiency L-infinity SLAM against BA-SLAM for rotation only camera motions, we measured the runtime of incremental rotation averaging and incremental BA, which are the fundamental optimisation routines for this problem. We used a window size equal to 10 and plotted the runtimes for the first loop of the Maptek dataset in Fig.~\ref{fig:rotonly} (in log scale). Rotation averaging is an order of magnitude faster than BA which indicates the superior efficiency of L-infinity SLAM over BA-SLAM for rotation only problems.



\section{CONCLUSIONS}
We presented L-infinity SLAM to be a simpler alternative to SLAM systems based on bundle adjustment. Driven by globally optimal quasi-convex optimisation, there is no need to maintain an accurate map and camera motions at key-frame rate as demanded by systems based on bundle adjustment. Instead, the online effort is devoted to efficiently estimating camera orientations through rotation averaging. To efficiently solve loop closure, we proposed a variant of the known rotation problem which incorporates relative translation directions to accurately solve camera drifts when optimising over a sample of feature tracks. Also, L-infinity SLAM is a simple and efficient alternative for applications requiring estimating slow motions or only rotational motions. We hope L-infinity SLAM can motivate future research on quasi-convex optimisation in the SLAM community.





\section*{ACKNOWLEDGEMENT}

This work was supported by ARC Grants LP140100946 and CE140100016.

\clearpage
\pagebreak

\bibliographystyle{IEEEtran}
\bibliography{ms}

\end{document}